\documentclass{article}

    \PassOptionsToPackage{numbers, compress}{natbib}

     \usepackage[final]{neurips_2019}

\usepackage[utf8]{inputenc} 
\usepackage[T1]{fontenc}    
\usepackage{hyperref}       
\usepackage{url}            
\usepackage{booktabs}       
\usepackage{amsfonts}       
\usepackage{nicefrac}       
\usepackage{microtype}      

\usepackage{amsmath}
\usepackage{amsfonts}
\usepackage{mathabx}
\usepackage{xcolor}
\usepackage[capitalize]{cleveref}
\usepackage{enumitem}
\usepackage{multirow}
\usepackage{graphicx}
\usepackage[mathscr]{euscript}
\usepackage{makecell}
\usepackage{soul}

\DeclareMathOperator*{\argmax}{arg\,max}

\let\vec\mathbf

\title{Natural Question Generation with Reinforcement Learning Based Graph-to-Sequence Model}

\author{
        Yu Chen\\
        Rensselaer Polytechnic Institute\\
        \texttt{cheny39@rpi.edu} 
    \And    
    Lingfei Wu\thanks{Corresponding author.} \\
    IBM Research\\
    \texttt{lwu@email.wm.edu} 
    \And
    Mohammed J. Zaki\\
    Rensselaer Polytechnic Institute\\
    \texttt{zaki@cs.rpi.edu}
    }%
        
\begin{document}

\maketitle

\begin{abstract}

Natural question generation (QG) aims to generate questions from a passage and an answer. In this paper, we propose a novel reinforcement learning (RL) based graph-to-sequence (Graph2Seq) model for QG. Our model consists of a Graph2Seq generator where a novel Bidirectional Gated Graph Neural Network is proposed to embed the passage, and a hybrid evaluator with a mixed objective combining both cross-entropy and RL losses to ensure the generation of syntactically and semantically valid text. The proposed model outperforms previous state-of-the-art methods by a large margin on the SQuAD dataset.

\end{abstract}

\vspace{-6mm}
\section{Introduction}
\vspace{-2mm}

Natural question generation (QG) is a dual task to question answering~\citep{chen2019bidirectional,chen2019graphflow}.
Given a passage \scalebox{0.95}{$X^p=\{x_1^p, x_2^p, ..., x_N^p\}$} and a target answer \scalebox{0.95}{$X^a=\{x_1^a, x_2^a, ..., x_L^a\}$}, the goal of QG is to generate the best question \scalebox{0.95}{$\hat{Y}=\{y_1, y_2, ..., y_T\}$} which maximizes the conditional likelihood \scalebox{0.95}{$\hat{Y} = \argmax_Y P(Y|X^p, X^a)$}.
Recent works on QG mostly formulate it as a sequence-to-sequence (Seq2Seq) learning problem~\citep{du2017learning,yao2018teaching,kumar2018automating}. 
However, these methods fail to utilize the rich text structure that could complement the simple word sequence.
Cross-entropy based sequence training has notorious limitations like exposure bias and inconsistency between train/test measurement~\citep{ranzato2015sequence,wu2016google,paulus2017deep}.
To tackle these limitations, some recent QG approaches~\citep{song2017unified,kumar2018framework} aim at directly optimizing evaluation metrics using Reinforcement Learning (RL)~\citep{williams1992simple}. However, they generally do not consider joint mixed objective functions with both syntactic and semantic constraints for guiding  text generation.

Early works on neural QG did not take into account the answer information when generating a question. Recent works~\citep{zhou2017neural,kim2018improving} have explored various means of utilizing the answers to make the generated questions more relevant.
However, they neglect potential semantic relations between the passage and answer, and thus fail to explicitly model the global interactions among them.

\begin{figure}[tb]
  \centering
    \includegraphics[keepaspectratio=true,scale=0.18]{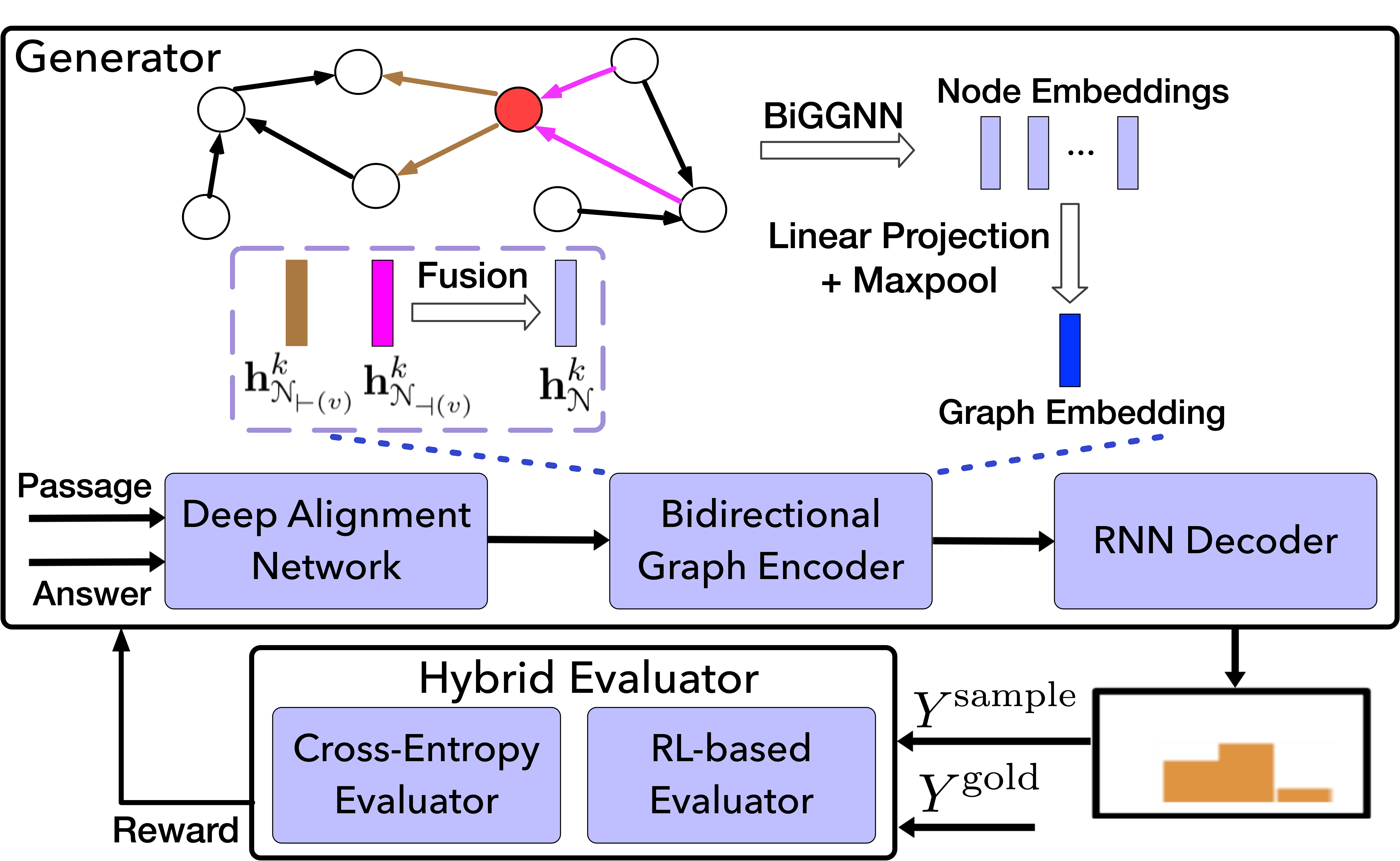}
  \caption{Overall architecture of the proposed model. Best viewed in color.}
  \label{fig:overall_arch}
\end{figure}

To address these aforementioned issues, as shown in \cref{fig:overall_arch},
we propose an RL based generator-evaluator architecture for QG, where the answer information is utilized by an effective Deep Alignment Network.
Our generator extends Gated Graph Neural Networks~\citep{li2015gated} by considering both incoming and outgoing edge information via a Bidirectional Gated Graph Neural Network (BiGGNN) for encoding the passage graph, and then outputs a question using an RNN-based decoder. Our hybrid evaluator is trained by optimizing a mixed objective function combining both cross-entropy loss and RL loss.
We also introduce an effective Deep Alignment Network for incorporating the answer information into the passage.
The proposed model is end-to-end trainable, and outperforms previous state-of-the-art methods by a great margin on the SQuAD dataset.

\vspace{-2mm}
\section{An RL-based generator-evaluator architecture}
\vspace{-2mm}
\textbf{Deep alignment network.}
Answer information is crucial for generating relevant and high quality questions from a passage.
 However, previous methods often neglect potential semantic relations between passage and answer words. 
 We thus propose a novel Deep Alignment Network (DAN) for effectively incorporating answer information into the passage by performing soft-alignment at both word-level and contextualized hidden state level.

Let \scalebox{0.97}{$\vec{X}^p \in \mathbb{R}^{F \times N}$} and  \scalebox{0.97}{$\widetilde{\vec{X}}^p \in \mathbb{R}^{\widetilde{F}_p \times N}$} denote two embeddings associated with passage text. Similarly, let \scalebox{0.97}{$\vec{X}^a \in \mathbb{R}^{F \times L}$} and  \scalebox{0.97}{$\widetilde{\vec{X}}^a \in \mathbb{R}^{\widetilde{F}_a \times L}$} denoted two embeddings associated with answer text. 
Formally, we define our soft-alignment function as
\scalebox{0.97}{
$\widetilde{\vec{H}}^p = \text{Align}(\vec{X}^p, \vec{X}^a, \widetilde{\vec{X}}^p, \widetilde{\vec{X}}^a) = \text{CAT}(\widetilde{\vec{X}}^p; \widetilde{\vec{X}}^a \boldsymbol{\beta}^T)$}
where \scalebox{0.97}{$\widetilde{\vec{H}}^p  \in \mathbb{R}^{(\widetilde{F}_p + \widetilde{F}_a) \times N}$} is the final passage embedding, CAT denotes concatenation, and $\boldsymbol{\beta}$ is a \scalebox{0.97}{$N \times L$} attention score matrix, computed by
\scalebox{0.97}{$\boldsymbol{\beta} \ \propto \ \text{exp} ( \text{ReLU}(\vec{W} \vec{X}^p)^T \text{ReLU}(\vec{W} \vec{X}^a))$}.
\scalebox{0.97}{$\vec{W} \in \mathbb{R}^{d \times F}$} is a trainable weight matrix, with $d$ being the hidden state size and \text{ReLU} is the rectified linear unit~\citep{nair2010rectified}.
We next introduce how we do soft-alignment at both word-level and contextualized hidden state level.

{\em Word-level alignment:}
In the word-level alignment stage, we first perform a soft-alignment between the passage and the answer based only on their pretrained GloVe embeddings and compute the final passage embeddings by \scalebox{0.97}{$\widetilde{\vec{H}}^p = \text{Align}(\vec{G}^p, \vec{G}^a, [\vec{G}^p; \vec{B}^p; \vec{L}^p], \vec{G}^a)$}, where
$\vec{G}^p$, $\vec{B}^p$, and $\vec{L}^p$ are the corresponding GloVe embedding~\citep{pennington2014glove}, BERT embedding~\citep{devlin2018bert}, and linguistic feature (i.e., case, NER and POS) embedding of the passage text, respectively. 
Then a bidirectional LSTM~\citep{hochreiter1997long} is applied to the passage embeddings \scalebox{0.97}{$\widetilde{\vec{H}}^p=\{\widetilde{\vec{h}}_i^p\}_{i=1}^N$} to obtain contextualized passage embeddings \scalebox{0.97}{$\widebar{\vec{H}}^p \in \mathbb{R}^{\widebar{F} \times N}$}. 
Similarly, on the answer side, we simply concatenate its GloVe embedding $\vec{G}^a$ with its BERT embedding $\vec{B}^a$.
Another BiLSTM is then applied to the above concatenated answer embedding sequence to obtain the contextualized answer embeddings \scalebox{0.97}{$\widebar{\vec{H}}^a \in \mathbb{R}^{\widebar{F} \times L}$}.

{\em Hidden-level alignment:}
In the hidden-level alignment stage, we perform another soft-alignment based on the contextualized passage and answer embeddings.
Similarly, we compute the aligned answer embedding, and concatenate it with the contextualized passage embedding to obtain the final passage embedding matrix \scalebox{0.97}{$\text{Align}([\vec{G}^p; \vec{B}^p; \widebar{\vec{H}}^p], [\vec{G}^a; \vec{B}^a;\widebar{\vec{H}}^a], \widebar{\vec{H}}^p, \widebar{\vec{H}}^a)$}.
Finally, we apply another BiLSTM to the above concatenated embedding to get a $\widebar{F} \times N$ passage embedding matrix $\vec{X}$.

\textbf{Bidirectional graph encoder.}
Existing methods have exploited RNNs to capture local dependencies among sequence words, which, however, neglects rich hidden structured information in text. GNNs provide a better approach to utilize the rich text structure and to model the global interactions among sequence words \cite{xu2018exploiting,xu2018sql,Subburathinam2019neural}. 
Therefore, we explore various ways of constructing a passage graph $\mathcal{G}$ containing each word as a node, and then apply GNNs to encode the passage graph.

{\em Passage graph construction:}
We explore syntax-based static graph construction.
For each sentence in a passage, we first get its dependency parse tree. We then connect neighboring dependency parse trees by connecting those nodes that are at a sentence boundary and next to each other in text.

We also explore semantics-aware dynamic graph construction, which consists of three steps: i) we compute a dense adjacency matrix $\vec{A}$ for the passage graph by applying self-attention to the passage embeddings $\widetilde{\vec{H}}^p$; ii) 
KNN-style graph sparsification is adopted to obtain a sparse adjacency matrix $\bar{\vec{A}}$; and iii) we apply softmax to $\bar{\vec{A}}$ to get two normalized adjacency matrices, namely, $\vec{A}^{\dashv}$ and $\vec{A}^{\vdash}$, for incoming and outgoing directions, respectively.
Please refer to~\cref{sec:dynamic_graph_construction} for more details.

{\em Bidirectional gated graph neural networks:}
Unlike \cite{xu2018graph2seq}, we propose a novel BiGGNN which is an extension to Gated Graph Neural Networks (GGNNs)~\citep{li2015gated}, to process the directed passage graph.
Node embeddings are initialized to the passage embeddings $\vec{X}$ returned by DAN.

At each hop of BiGGNN, for every node, we apply an aggregation function which takes as input a set of incoming (or outgoing) neighboring node vectors and outputs a backward (or forward) aggregation vector, denoted as $\vec{h}^k_{\EuScript{N}_{\dashv(v)}}$ (or $\vec{h}^k_{\EuScript{N}_{\vdash(v)}}$).
For the syntax-based static graph, the backward (or forward) aggregation vector is computed as the average of all incoming (or outgoing) neighboring node vectors plus itself.
For the semantics-aware dynamic graph, we compute a weighted average for aggregation where the weights come from the adjacency matrices $\vec{A}^{\dashv}$ and $\vec{A}^{\vdash}$.

Unlike \citep{xu2018graph2seq} that learns separate node embeddings for both directions independently,
we fuse the information aggregated in two directions at each hop, defined as, 
\scalebox{0.97}{$\vec{h}^k_{\EuScript{N}}=\text{Fuse}(\vec{h}^k_{\EuScript{N}_{\dashv(v)}}, \vec{h}^k_{\EuScript{N}_{\vdash(v)}})$}.
And the fusion function is designed as a gated sum of two information sources, defined as, \scalebox{0.97}{$\text{Fuse}(\vec{a}, \vec{b}) = \vec{z} * \vec{a} + (1-\vec{z}) * \vec{b}$} where
\scalebox{0.97}{$\vec{z} = \sigma(\vec{W}_{\!z} [\vec{a}; \vec{b}; \vec{a}*\vec{b}; \vec{a}-\vec{b}]+\vec{b}_z)$} with $\sigma$ being a sigmoid function and $\vec{z}$ being a gating vector.
Finally, a Gated Recurrent Unit (GRU)~\citep{cho2014learning} is used to update the node embeddings by incorporating the aggregation information, defined as, \scalebox{0.97}{$\vec{h}^{k}_v=\text{GRU}(\vec{h}^{k-1}_v, \vec{h}^k_{\EuScript{N}})$}.

After $n$ hops of GNN computation, where $n$ is a hyperparameter,
we obtain the final state embedding $\vec{h}^{n}_v$ for node $v$.
To compute the graph-level embedding, we first apply a linear projection to the node embeddings, and then apply max-pooling over all node embeddings to get a $d$-dim vector $\vec{h}^\mathcal{G}$.
The decoder takes the graph-level embedding $\vec{h}^\mathcal{G}$ followed by two separate fully-connected layers as initial hidden states (i.e., $\vec{c}_0$ and $\vec{s}_0$) and the node embeddings $\{\vec{h}^{n}_v, \forall v \in \mathcal{G}\}$ as the attention memory.
Our decoder closely follows~\citep{see2017get}. We refer the readers to~\cref{sec:rnn_decoder} for more details.

\textbf{Hybrid evaluator.}
Some recent QG approaches~\citep{song2017unified,kumar2018framework} directly optimize evaluation metrics using REINFORCE to overcome the loss mismatch issue with cross-entropy based sequence training. However, they often fail to generate semantically meaningful and syntactically coherent text. 
To address these issues, we present a hybrid evaluator with a mixed objective combining both cross-entropy and RL losses so as to ensure the generation of syntactically and semantically valid text.

For the RL part, we adopt the self-critical sequence training (SCST) algorithm~\citep{rennie2017self} to directly optimize evaluation metrics.
In SCST, at each training iteration,
the model generates two output sequences: the sampled output $Y^s$ produced by multinomial sampling, 
and the baseline output $\hat{Y}$ obtained by greedy search. 
We define $r(Y)$ as the reward of an output sequence $Y$, computed by comparing it to corresponding ground-truth sequence $Y^*$ with some reward metrics.
The loss function is defined as
$\mathcal{L}_{rl}=(r(\hat{Y})-r(Y^s)) \sum_{t}{\log{} P(y_t^s|X, y^s_{<t})}$.
As we can see, if the sampled output has a higher reward than the baseline one, we maximize its likelihood, and vice versa.

We use one of our evaluation metrics, BLEU-4, as our reward function $f_{\text{eval}}$, which lets us directly optimize the model towards the evaluation metrics.
One drawback of some evaluation metrics like BLEU is that they do not measure meaning, but only reward systems for n-grams that have exact matches in the reference system.
To make our reward function more effective and robust, 
following~\citep{gong2019reinforcement}, we additionally use word mover’s distance (WMD)~\citep{kusner2015word} as a semantic reward function $f_{\text{sem}}$.
We define the final reward function as $r(Y) = f_{\text{eval}}(Y, Y^*) + \alpha f_{\text{sem}}(Y, Y^*)$ where $\alpha$ is a scalar.

We train our model in two stages. 
In the first state, we train the model using regular cross-entropy loss. And in the second stage, we fine-tune the model by optimizing a mixed objective function combining both cross-entropy loss and RL loss.
During the testing phase, we use beam search to generate final predictions.
Further details of the training strategy can be found in~\cref{sec:training_details}.

\vspace{-3mm}
\section{Experiments}
\vspace{-3mm}

In this section, we evaluate our proposed model against state-of-the-art methods on the SQuAD dataset~\citep{rajpurkar2016squad}.  
The baseline methods in our experiments include
\text{SeqCopyNet}~\citep{zhou2018sequential},
\text{NQG++}~\citep{zhou2017neural}, 
\text{MPQG+R}~\citep{song2017unified},
\text{Answer-focused Position-aware model}~\citep{sun2018answer},
\text{s2sa-at-mp-gsa}~\citep{zhao2018paragraph},
\text{ASs2s}~\citep{kim2018improving} and \text{CGC-QG}~\citep{liu2019learning}.
Note that experiments on baselines followed by \text{*} are conducted using released source code. 
Detailed description of the baselines is provided in~\cref{sec:baselines}.
For fair comparison with baselines, we do experiments on both SQuAD split-1~\citep{song2018leveraging} and  split-2~\citep{zhou2017neural}.
For model settings and sensitivity analysis of hyperparameters, please refer to~\cref{sec:model_settings} and~\cref{sec:sensitivity_analysis}.
The implementation of our model will be made publicly available at \url{https://github.com/hugochan/RL-based-Graph2Seq-for-NQG}.

Following previous works, we use BLEU-4~\citep{papineni2002bleu}, METEOR~\citep{banerjee2005meteor} and ROUGE-L~\citep{lin2004rouge} as our evaluation metrics. 
Note that we only report BLEU-4 on split-2 since most of the baselines only report this result.
Besides automatic evaluation metrics, we also conduct human evaluation on split-2.
Further details on human evaluation can be found in~\cref{sec:human_evaluation}.


\begin{table}[tbh]
\caption{Automatic evaluation results on the SQuAD test set.}
\vspace{-3mm}
\label{table:squad_results}
\begin{center}
\scalebox{0.96}{
\begin{tabular}{lllllll}
\hline
 \multirow{2}{*}{Methods}& \vline &  &Split-1&& \vline & Split-2\\
    & \vline &  BLEU-4 & METEOR & ROUGE-L & \vline & BLEU-4 \\
  \hline
  \hline
\text{SeqCopyNet} & \vline & \quad \textrm{--}&\quad \textrm{--}& \quad \textrm{--}&\vline &13.02\\
\text{NQG++} & \vline & \quad \textrm{--}&\quad \textrm{--}& \quad \textrm{--}&\vline &13.29\\
 \text{MPQG+R*} & \vline & 14.39 &18.99&42.46 &\vline&14.71\\
 \text{Answer-focused Position-aware model} & \vline & \quad \textrm{--}&\quad \textrm{--}& \quad \textrm{--}&\vline &15.64\\
 \text{s2sa-at-mp-gsa} & \vline &15.32& 19.29&  43.91&\vline &15.82\\
 \text{ASs2s} & \vline & 16.20 & 19.92 & 43.96 & \vline&16.17\\
 \text{CGC-QG} & \vline &\quad \textrm{--} &\quad \textrm{--}&\quad \textrm{--}&\vline& 17.55\\
  \hline
\text{G2S}$_{dyn}$\text{+BERT+RL} & \vline &17.55 &21.42&45.59&\vline&18.06\\
\text{G2S}$_{sta}$\text{+BERT+RL} & \vline &\textbf{17.94} &\textbf{21.76}&\textbf{46.02}&\vline&\textbf{18.30}\\
 \hline
\end{tabular}
}
\end{center}
\vspace{-4mm}
\end{table}

\textbf{Experimental results and analysis.}
\cref{table:squad_results} shows the automatic evaluation results comparing against all baselines. 
First of all, our full model \text{G2S}$_{sta}$\text{+BERT+RL} outperforms previous state-of-the-art methods by a great margin.
Compared to previous methods like \text{CGC-QG}~\citep{liu2019learning} and \text{ASs2s}~\citep{kim2018improving} that rely on many heuristic rules and ad-hoc strategies,
our model does not rely on any of these hand-crafted rules and ad-hoc strategies.


\begin{table}[!htb]
\caption{Ablation study on the SQuAD split-2 test set.}
\label{table:squad_ablation_results}
\centering
\scalebox{0.96}{
\begin{tabular}{lllllll}
\hline
 Methods& \vline & BLEU-4 & \vline & Methods& \vline & BLEU-4 \\
  \hline
\text{G2S}$_{dyn}$\text{+BERT+RL} & \vline &18.06 & \vline
& \text{G2S}$_{dyn}$ & \vline &16.81\\
 
 \text{G2S}$_{sta}$\text{+BERT+RL} & \vline &18.30 & \vline
& \text{G2S}$_{sta}$ & \vline &16.96\\
 
\text{G2S}$_{sta}$\text{+BERT-fixed+RL} & \vline &18.20& \vline 
& \text{G2S}$_{dyn}$ \text{w/o DAN} & \vline &12.58\\ 

\text{G2S}$_{dyn}$\text{+BERT} & \vline &17.56& \vline
& \text{G2S}$_{sta}$ \text{w/o DAN} & \vline &12.62\\

\text{G2S}$_{sta}$\text{+BERT} & \vline & 18.02& \vline 
& \text{G2S}$_{sta}$ \text{w/o BiGGNN}, w/ \text{Seq2Seq} & \vline & 16.14\\

  \text{G2S}$_{sta}$\text{+BERT-fixed} & \vline & 17.86& \vline
& \text{G2S}$_{sta}$ \text{w/o BiGGNN, w/ GCN} & \vline & 14.47 \\
  
  \text{G2S}$_{dyn}$\text{+RL}  &\vline &17.18& \vline 
  & \text{G2S}$_{sta}$ \text{w/ GGNN-forward} & \vline &16.53\\
  
  \text{G2S}$_{sta}$\text{+RL} & \vline & 17.49& \vline
  & \text{G2S}$_{sta}$ \text{w/ GGNN-backward} & \vline &16.75\\

    \hline
\end{tabular}
}
\end{table}

We also perform ablation study on the impact of different components on the SQuAD split-2 test set, as shown in~\cref{table:squad_ablation_results}.
For complete results on ablation study, please refer to~\cref{sec:complete_ablation_results}.
By turning off DAN, the BLEU-4 score of \text{G2S}$_{sta}$ (similarly for \text{G2S}$_{dyn}$) dramatically drops from $16.96\%$ to $12.62\%$, which shows the effectiveness of DAN.
We can see the advantages of Graph2Seq learning over Seq2Seq learning by comparing the performance between \text{G2S}$_{sta}$ and \text{Seq2Seq}.
Fine-tuning the model using REINFORCE can further improve the model performance, which shows the benefits of directly optimizing evaluation metrics.
We also find that BERT has a considerable impact on the performance.
Lastly, we find that static graph construction slightly outperforms dynamic graph construction.
We refer the readers to~\cref{sec:case_study} for a case study of different ablated systems.

\section{Conclusion}

 We proposed a novel RL based Graph2Seq model for QG, where the answer information is utilized by an effective Deep Alignment Network and a novel bidirectional GNN is proposed to process the directed passage graph.
 Our two-stage training strategy benefits from both cross-entropy based and REINFORCE based sequence training.
 We also explore both static and dynamic approaches for constructing graphs when applying GNNs to textual data.
On the SQuAD dataset, our model outperforms previous state-of-the-art methods by a wide margin.
In the future, we would like to investigate more effective ways of automatically learning graph structures from free-form text.

\subsubsection*{Acknowledgments}

This work is supported by IBM Research AI through the IBM AI Horizons Network. 
We thank the human evaluators who evaluated our system.
We thank the anonymous reviewers for their feedback.






\bibliography{gnn_qg_neurips2019}
\bibliographystyle{abbrv}

\appendix

\section{Details on dynamic graph construction}\label{sec:dynamic_graph_construction}

We dynamically build a directed and weighted graph to model semantic relationships among passage words. We make the process of building such a graph depend on not only the passage, but also on the answer. 
The graph construction procedure consists of three steps: i) we compute a dense adjacency matrix $\vec{A}$ for the passage graph by applying self-attention to the word-level passage embeddings $\widetilde{\vec{H}}^p$; ii) 
a KNN-style graph sparsification strategy is adopted to obtain a sparse adjacency matrix $\bar{\vec{A}}$, where we only keep the $K$ nearest neighbors (including itself) as well as the associated attention scores (i.e., the remaining attentions scores are masked off) for each node; and iii) we apply softmax to $\bar{\vec{A}}$ to get two normalized adjacency matrices, namely, $\vec{A}^{\dashv}$ and $\vec{A}^{\vdash}$, for incoming and outgoing directions, respectively.
\begin{equation}\label{eq:dyn_graph}
\vec{A} = \text{ReLU}(\vec{U} \widetilde{\vec{H}}^p)^T\ \text{ReLU}(\vec{U} \widetilde{\vec{H}}^p)\quad
\bar{\vec{A}} = \text{KNN}(\vec{A})\quad
\vec{A}^{\dashv}, \vec{A}^{\vdash} = \text{softmax}(\{\bar{\vec{A}}, \bar{\vec{A}}^T\})
\end{equation}
where $\vec{U}$ is a $d \times (\widetilde{F}_p + \widetilde{F}_a)$ trainable weight matrix.
Note that the supervision signal is able to back-propagate through the KNN-style graph sparsification operation since the $K$ nearest attention scores are kept.

\section{Details on the RNN decoder}\label{sec:rnn_decoder}
On the decoder side, we adopt an attention-based~\citep{bahdanau2014neural,luong2015effective} LSTM decoder with copy~\citep{vinyals2015pointer,gu2016incorporating} and coverage mechanisms~\citep{tu2016modeling}.
At each decoding step $t$,
an attention mechanism learns to attend to the most relevant words in the input sequence, and computes a context vector $\vec{h}^*_t$ based on the current decoding state $\vec{s}_t$, the current coverage vector $\vec{c}^t$ and the attention memory.
In addition, the generation probability $p_\text{gen} \in [0, 1]$ is calculated from the context vector $\vec{h}^*_t$, the decoder state $\vec{s}_t$ and the decoder input $y_{t-1}$.
Next, $p_\text{gen}$ is used as a soft switch to choose between generating a word from the vocabulary, or copying a word from the input sequence.
We dynamically maintain an extended vocabulary which is the union of the usual vocabulary and all words appearing in a batch of source examples (i.e., passages and answers).
Finally, in order to encourage the decoder to utilize the diverse components of the input sequence, a coverage mechanism is applied.
At each step, we maintain a coverage vector $\vec{c}^t$, which is the sum of attention distributions over all previous decoder time steps.
A coverage loss is also computed to penalize repeatedly attending to the same locations of the input sequence.

\section{Training details}\label{sec:training_details}
We train our model in two stages. 
In the first state, we train the model using regular cross-entropy loss, defined as,
$\mathcal{L}_{lm} = \sum_{t}{}-\log{} P(y_t^*|X, y^*_{<t}) + \lambda\ \text{covloss}_t$ 
where $y_t^*$ is the word at the $t$-th position of the ground-truth output sequence and $\text{covloss}_t$ is the coverage loss defined as $\sum_i{min(a_i^t, c_i^t)}$, with $a_i^t$ being the $i$-th element of the attention vector over the input sequence at time step $t$.
Scheduled teacher forcing~\citep{bengio2015scheduled} is adopted to alleviate the exposure bias problem. 
In the second stage, we fine-tune the model by optimizing a mixed objective function combining both cross-entropy loss and RL loss, defined as, $
\mathcal{L}= \gamma \mathcal{L}_{rl} + (1 - \gamma) \mathcal{L}_{lm}$ where $\gamma$ is a scaling factor controling the trade-off between cross-entropy loss and RL loss.
A similar mixed-objective learning function has been used by~\citep{wu2016google,paulus2017deep} for machine translation and text summarization.

\section{Details on baseline methods}\label{sec:baselines}

\noindent\textbf{SeqCopyNet}~\citep{zhou2018sequential}
proposed an extension to the copy mechanism which learns to copy not only single words but also sequences from the input sentence.

\noindent\textbf{NQG++}~\citep{zhou2017neural}
proposed an attention-based Seq2Seq model equipped with copy mechanism and a feature-rich encoder to encode answer position, POS and NER tag information.

\noindent\textbf{MPQG+R}~\citep{song2017unified}
proposed an RL-based Seq2Seq model with a multi-perspective matching encoder to incorporate answer information.
Copy and coverage mechanisms are applied.

\noindent\textbf{Answer-focused Position-aware model}~\citep{sun2018answer} 
consists of an answer-focused component which generates an interrogative word matching the answer type, and a position-aware component which is aware of the position of the context words when generating a question by modeling the relative distance between the context words and the answer.

\noindent\textbf{s2sa-at-mp-gsa}~\citep{zhao2018paragraph}
proposed a model which contains a gated attention encoder and a maxout pointer decoder to tackle the challenges of processing long input sequences. For fair comparison, we report the results of the sentence-level version of their model to match with our settings.

\noindent\textbf{ASs2s}~\citep{kim2018improving}
proposed an answer-separated Seq2Seq model which treats the passage and the answer separately.

\noindent\textbf{CGC-QG}~\citep{liu2019learning}
proposed a multi-task learning framework to guide the model to learn the accurate boundaries between copying and generation.

\section{Model settings}\label{sec:model_settings}
We keep and fix the 300-dim GloVe vectors for the most frequent 70,000 words in the training set.
We compute the 1024-dim BERT embeddings on the fly for each word in text using a (trainable) weighted sum of all BERT layer outputs.
The embedding sizes of case, POS and NER tags are set to 3, 12 and 8, respectively.
We set the hidden state size of BiLSTM to 150 so that the concatenated state size for both directions is 300.
The size of all other hidden layers is set to 300.
We apply a variational dropout~\citep{kingma2015variational} rate of 0.4 after word embedding layers and 0.3 after RNN layers. 
We set the neighborhood size to 10 for dynamic graph construction.
The number of GNN hops is set to 3.
During training, 
in each epoch, we set the initial teacher forcing probability to 0.75 and exponentially increase it to $0.75 * 0.9999^i$ where $i$ is the training step.
We set $\alpha$ in the reward function to 0.1,
$\gamma$ in the mixed loss function to 0.99,
and the coverage loss ratio $\lambda$ to 0.4.
We use Adam \citep{kingma2014adam} as the optimizer and the learning rate is set to 0.001 in the pretraining stage and 0.00001 in the fine-tuning stage.
We reduce the learning rate by a factor of 0.5 if the validation BLEU-4 score stops improving for three epochs. We stop the training when no improvement is seen for 10 epochs.
We clip the gradient at length 10.
The batch size is set to 60 and 50 on data split-1 and split-2, respectively.
The beam search width is set to 5.
All hyperparameters are tuned on the development set.

\section{Sensitivity analysis of hyperparameters}\label{sec:sensitivity_analysis}

\begin{figure}[!htb]
  \centering
    \includegraphics[keepaspectratio=true,scale=0.3]{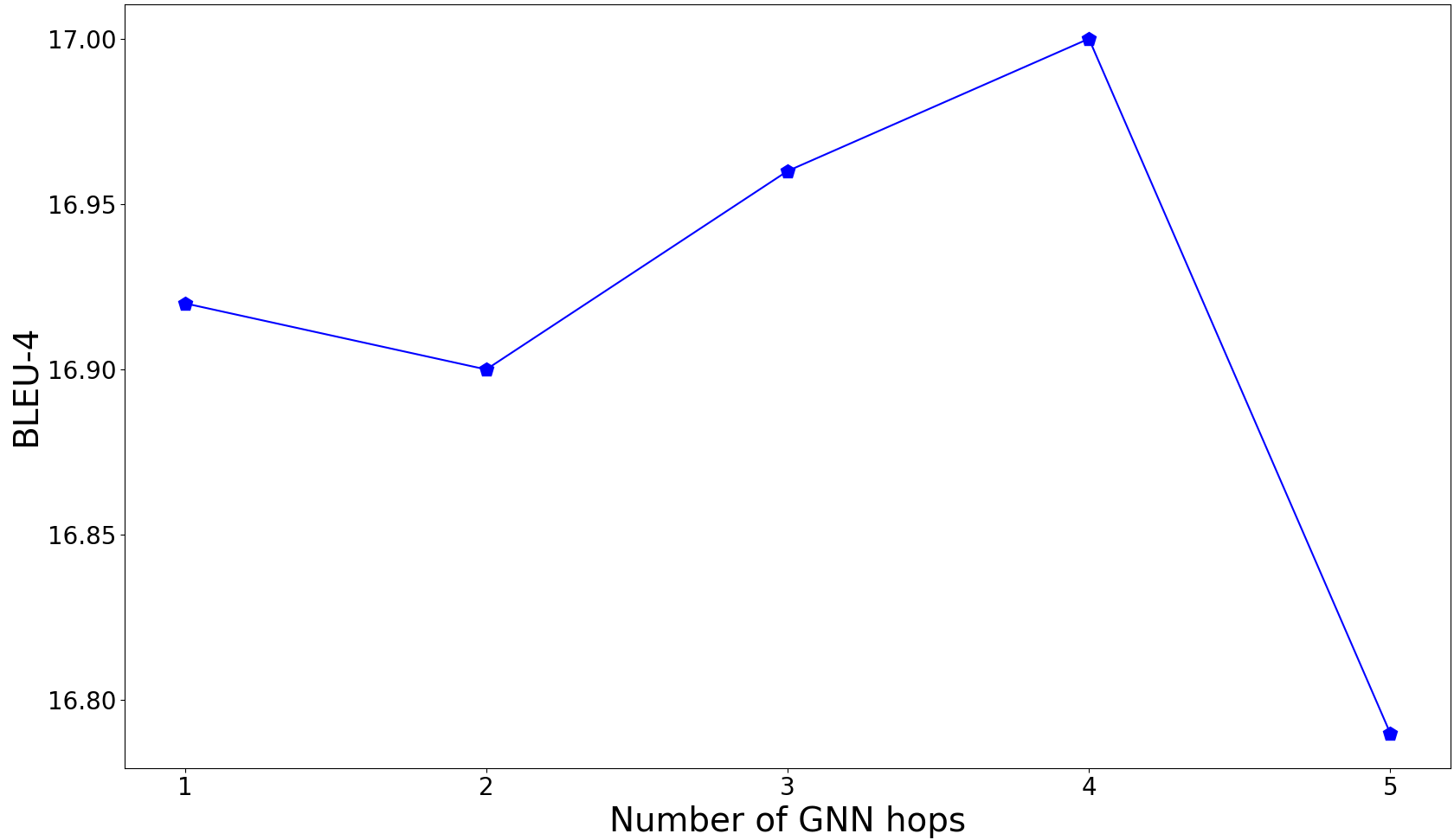}
  \caption{Effect of the number of GNN hops.}
  \label{fig:effect_gnn_hops}
\end{figure}

We study the effect of the number of GNN hops on our model performance. We conduct experiments of the \text{G2S}$_{sta}$ model on the SQuAD split-2 data by varying the number of GNN hops.
\cref{fig:effect_gnn_hops} shows that our model is not very sensitive to the number of GNN hops and can achieve reasonable good results with various number of hops.

\section{Details on human evaluation}\label{sec:human_evaluation}
We conducted a small-scale (i.e., 50 random examples per system) human evaluation on the split-2 data.
We asked 5 human evaluators to give feedback on the quality of questions generated by a set of anonymised competing systems.
In each example, given a triple containing a source passage, a target answer and an anonymised system output, they were asked to rate the quality of the system output by answering the following three questions: i) is this generated question syntactically correct? ii) is this generated question semantically correct? and iii) is this generated question relevant to the passage?
For each evaluation question, the rating scale is from 1 to 5, where a higher score means better quality (i.e., 1: Poor, 2: Marginal, 3: Acceptable, 4: Good, 5: Excellent).
Responses from all evaluators were collected and averaged.

As shown in~\cref{table:squad_human_evaluation_results}, we conducted a human evaluation study to assess the quality of the questions generated by our model, the baseline method MPQG+R, and the ground-truth data in terms of syntax, semantics and relevance metrics. 
We can see that our best performing model achieves good results even compared to the ground-truth, and outperforms the strong baseline method MPQG+R.
Our error analysis shows that main syntactic error occurs in repeated/unknown words in generated questions. Further, the slightly lower quality on semantics also impacts the relevance.

\begin{table}[!htb]
\caption{Human evaluation results on the SQuAD split-2 test set.}
\label{table:squad_human_evaluation_results}
\centering
\begin{tabular}{lllll}
\hline
  Methods & \vline & Syntactically correct (\%) & Semantically correct (\%) & Relevant (\%)\\
  \hline
  \hline
  \text{MPQG+R*} & \vline &4.34 &4.01 & 3.21\\
 \text{G2S}$_{sta}$\text{+BERT+RL} & \vline &4.41 &4.31 & 3.79\\
  Ground-truth & \vline & \textbf{4.74} &\textbf{4.74} & \textbf{4.25}\\
 \hline
\end{tabular}
\end{table}

\section{Complete results on Ablation Study}\label{sec:complete_ablation_results}

\begin{table}[!htb]
\caption{Ablation study on the SQuAD split-2 test set.}
\label{app:table:squad_ablation_results}
\centering
\begin{tabular}{lllllll}
\hline
 Methods& \vline &   BLEU-4 & \vline & Methods& \vline &   BLEU-4\\
  \hline
\text{G2S}$_{dyn}$\text{+BERT+RL} & \vline &18.06 & \vline
 & \text{G2S}$_{dyn}$ \text{w/o feat} & \vline &16.51\\
 
 \text{G2S}$_{sta}$\text{+BERT+RL} & \vline &18.30 & \vline
 & \text{G2S}$_{sta}$ \text{w/o feat} & \vline &16.65\\
 
\text{G2S}$_{sta}$\text{+BERT-fixed+RL} & \vline &18.20& \vline 
& \text{G2S}$_{dyn}$ \text{w/o DAN} & \vline &12.58\\ 

\text{G2S}$_{dyn}$\text{+BERT} & \vline &17.56& \vline
& \text{G2S}$_{sta}$ \text{w/o DAN} & \vline &12.62\\

\text{G2S}$_{sta}$\text{+BERT} & \vline & 18.02& \vline 
& \text{G2S}$_{sta}$ \text{w/ DAN-word only} & \vline & 15.92\\

  \text{G2S}$_{sta}$\text{+BERT-fixed} & \vline & 17.86& \vline
  & \text{G2S}$_{sta}$ \text{w/ DAN-hidden only} & \vline & 16.07\\  
  
  \text{G2S}$_{dyn}$\text{+RL}  &\vline &17.18& \vline 
  & \text{G2S}$_{sta}$ \text{w/ GGNN-forward} & \vline &16.53\\
  
  \text{G2S}$_{sta}$\text{+RL} & \vline & 17.49& \vline
  & \text{G2S}$_{sta}$ \text{w/ GGNN-backward} & \vline &16.75\\
  
  \text{G2S}$_{dyn}$ & \vline &16.81 &\vline
  & \text{G2S}$_{sta}$ \text{w/o BiGGNN}, w/ \text{Seq2Seq} & \vline & 16.14\\
  
  \text{G2S}$_{sta}$ & \vline &16.96 & \vline
   & \text{G2S}$_{sta}$ \text{w/o BiGGNN, w/ GCN} & \vline & 14.47 \\
    \hline
\end{tabular}
\end{table}

We perform the comprehensive ablation study to systematically assess the impact of different model components (e.g., BERT, RL, DAN, BiGGNN, FEAT, DAN-word, and DAN-hidden) for two proposed full model variants (static vs dynamic) on the SQuAD split-2 test set. Our experimental results confirmed that every component in our proposed model makes the contribution to the overall performance.

\section{Case study of ablated systems}\label{sec:case_study}

In \cref{table:squad_qg_examples}, we further show a few examples that illustrate the quality of generated text given a passage under different ablated systems. 
As we can see, incorporating answer information helps the model identify the answer type of the question to be generated, and thus makes the generated questions more relevant and specific.
Also, we find our Graph2Seq model can generate more complete and valid questions compared to the Seq2Seq baseline. We think it is because a Graph2Seq model is able to exploit the rich text structure information better than a Seq2Seq model. 
Lastly, it shows that fine-tuning the model using REINFORCE can improve the quality of the generated questions.

\begin{table}[!htb]
\caption{Generated questions on SQuAD split-2 test set. Target answers are underlined.}
\label{table:squad_qg_examples}
\centering
\begin{tabular}{l}
\hline
\textbf{Passage:} for the successful execution of a project , \underline{effective planning} is essential .\\

\textbf{Gold:} what is essential for the successful execution of a project ?\\


\textbf{G2S}$_{sta}$ \textbf{w/o BiGGNN (Seq2Seq):} what type of planning is essential for the project ?\\

\textbf{\text{G2S}$_{sta}$ \text{w/o DAN.}:}
what type of planning is essential for the successful execution of a project ?\\

\textbf{\text{G2S}$_{sta}$:} what is essential for the successful execution of a project ?\\

\textbf{\text{G2S}$_{sta}$\text{+BERT}:}
what is essential for the successful execution of a project ?\\

\textbf{\text{G2S}$_{sta}$\text{+BERT+RL}:}
what is essential for the successful execution of a project ?\\

\textbf{\text{G2S}$_{dyn}$\text{+BERT+RL}:}
what is essential for the successful execution of a project ?\\

\hline
\hline

\textbf{Passage:} the church operates \underline{three hundred sixty schools and institutions overseas} .\\

\textbf{Gold:} how many schools and institutions does the  church operate overseas ?     
\\


\textbf{G2S}$_{sta}$ \textbf{w/o BiGGNN (Seq2Seq):} how many schools does the church have ?
\\

\textbf{\text{G2S}$_{sta}$ \text{w/o DAN.}:}
how many schools does the church have ?\\

\textbf{\text{G2S}$_{sta}$:} how many schools and institutions does the church have ?\\

\textbf{\text{G2S}$_{sta}$\text{+BERT}:}
how many schools and institutions does the church have ?\\

\textbf{\text{G2S}$_{sta}$\text{+BERT+RL}:}
how many schools and institutions does the church operate ?\\

\textbf{\text{G2S}$_{dyn}$\text{+BERT+RL}:}
how many schools does the church operate ?\\
\hline
\end{tabular}
\end{table}

\end{document}